\pdfoutput=1
\documentclass{bmvc2k}
\usepackage[table]{xcolor}

\title{CAPE-JEPA — Cross-modal Adversarial Probabilistic Embedding JEPA}

\addauthor{Aniket Junghare}{aniketjunghare.iitb@gmail.com}{1}
\addauthor{Biplab Banerjee}{getbiplab@gmail.com}{1}
\runninghead{Junghare \& Banerjee}{CAPE-JEPA}
\addinstitution{
 Indian Institute of Technology Bombay\\
 Mumbai, India
}

\def\eg{\emph{e.g}\bmvaOneDot}

\usepackage{amsmath}
\usepackage{amssymb} 
\usepackage{booktabs}
\usepackage{multirow}
\usepackage{pifont}
\usepackage{float}

\usepackage{enumitem}
\usepackage{colortbl}
\usepackage{wrapfig}
\usepackage{caption}   
\newcommand{\cmark}{\ding{51}}
\newcommand{\xmark}{\ding{55}}

\definecolor{capblue}{RGB}{219,234,254}
\definecolor{capgreen}{RGB}{220,252,231}
\definecolor{capred}{RGB}{254,226,226}
\definecolor{capheader}{RGB}{241,245,249}
\definecolor{benblue}{RGB}{219,234,254}
\definecolor{fmowgreen}{RGB}{220,252,231}
\definecolor{unimodal}{RGB}{241,245,249}
\definecolor{crossmodal}{RGB}{254,243,199}
\definecolor{maerow}{RGB}{254,226,226}
\definecolor{crossrow}{RGB}{220,252,231}
\definecolor{jeparow}{RGB}{219,234,254}
\definecolor{ourrow}{RGB}{134,239,172}

\begin{document}

\maketitle

\begin{abstract}
Cross-modal remote sensing image retrieval (RS-CMIR) requires representations
that are simultaneously modality-invariant and semantically discriminative,
despite the profound physical gap between heterogeneous sensors such as
Synthetic Aperture Radar (SAR) and multispectral imagery. Recent JEPA-based Predictive
self-supervised approaches such as X-JEPA \cite{Choudhury_2026_WACV} formulate cross-modal alignment
as a deterministic latent forecasting task, implicitly assuming uniform
patch predictability and leaving cross-modal distributional alignment
unaddressed. We propose CAPE-JEPA, a probabilistic cross-modal Joint Embedding
Predictive Architecture with three contributions: (i) a cross-attentional
Gaussian predictor outputting per-patch parameters $(\mu, \sigma^{2})$
conditioned on opposite-modality context, trained via negative log-likelihood
to model inherent cross-sensor prediction uncertainty; (ii) adversarial
distributional regularization combining unimodal isotropic Gaussian regularization
with a novel cross-modal adversarial variant that aligns modality distributions
via worst-case Epps-Pulley characteristic function tests without explicit sample
pairing; and (iii) in-training contrastive learning with attention-pooled heads
for scene-level discriminative grounding. Evaluated on BigEarthNet-14K (BEN-14K)
and fMoW across SAR, multispectral, and RGB modalities, CAPE-JEPA surpasses
X-JEPA by up to \textbf{13.74 points} on BEN-14K and \textbf{15.59 points}
on fMoW, outperforming all baselines including SkySense and CROMA across
both unimodal and cross-modal directions while using fewer parameters
(407.30M vs.\ 432.72M).
\end{abstract}

\vspace{-10pt}
\section{Introduction}
\label{sec:intro}

The proliferation of multi-sensor satellite constellations has produced vast,
heterogeneous remote sensing (RS) archives spanning Synthetic Aperture Radar
(SAR), multispectral, and high-resolution optical
imagery~\cite{guo2024skysense,Torres2012TheSM,drusch2012sentinel}. Cross-modal remote sensing image retrieval (RS-CMIR) aims to retrieve semantically 
relevant images across sensor modalities~\cite{Choudhury_2026_WACV, fuller2023croma}, and is central to applications such as disaster response, land-use monitoring, and environmental analysis~\cite{guo2024skysense,sumbul2019bigearthnet}. The task is
fundamentally challenging: SAR measures microwave backscatter sensitive to
surface texture and dielectric properties, while multispectral sensors capture
reflected solar radiation across discrete spectral
bands~\cite{ulaby2014microwave,jensen2009remote}. This physical divergence means
semantically identical scenes appear visually uncorrelated across modalities,
rendering pixel-level comparison ineffective and demanding representations that
are simultaneously modality-invariant and semantically
discriminative~\cite{Choudhury_2026_WACV,liang2022mind}. In disaster response, fusing SAR's all-weather structural sensitivity with multispectral 
spectral richness is critical. Yet no retrieval framework has closed the distributional gap between these sensors in a principled, uncertainty-aware manner.

Self-supervised learning (SSL) has emerged as the dominant approach for
cross-modal alignment~\cite{wang2022self,fuller2023croma}, bypassing the
need for manually annotated paired data across heterogeneous RS sensor
pairs~\cite{fuller2023croma,Choudhury_2026_WACV}. Contrastive methods~\cite{chen2020simple,he2020momentum}
learn representations by pulling together views of the same scene while pushing
apart negatives, but suffer from a structural false-negative problem in RS:
visually dissimilar scenes from different sensors routinely share semantic
categories (\eg, flooded cropland in SAR and dry cropland in multispectral belong
to the same land-cover class), causing contrastive objectives to push apart
semantically equivalent embeddings and destabilise the learned
space~\cite{Choudhury_2026_WACV,wang2022self}. Masked image modelling (MIM) 
approaches~\cite{he2022masked,cong2022satmae} sidestep this by reconstructing masked pixel regions without requiring negative pairs. However, pixel-space reconstruction rewards modality-specific appearance cues such as speckle statistics, spectral profiles, and resolution artefacts, which must be abstracted 
away for cross-modal retrieval~\cite{Choudhury_2026_WACV,fuller2023croma}.

\begin{wrapfigure}{r}{0.50\columnwidth}
\vspace{-14pt}
\centering
\setlength{\abovecaptionskip}{4pt}
\setlength{\belowcaptionskip}{6pt}
\setlength{\tabcolsep}{4pt}
\renewcommand{\arraystretch}{1.15}
\resizebox{0.48\columnwidth}{!}{%
\begin{tabular}{lccccc}
\toprule
\rowcolor{capheader}
\textbf{Method} & \rotatebox{60}{\textbf{PP}}
                & \rotatebox{60}{\textbf{DA}}
                & \rotatebox{60}{\textbf{DG}}
                & \rotatebox{60}{\textbf{LS}}
                & \rotatebox{60}{\textbf{CM}} \\
\midrule
\rowcolor{capred!40}
MAE~\cite{he2022masked}
  & \xmark & \xmark & \xmark & \xmark & \xmark \\
\rowcolor{capred!20}
I-JEPA~\cite{assran2023self}
  & \xmark & \xmark & \xmark & \cmark & \xmark \\
\rowcolor{capred!20}
X-JEPA~\cite{Choudhury_2026_WACV}
  & \xmark & \xmark & \xmark & \cmark & \cmark \\
\midrule
\rowcolor{capgreen!60}
\textbf{CAPE-JEPA (ours)}
  & \cmark & \cmark & \cmark & \cmark & \cmark \\
\bottomrule
\end{tabular}%
}
\vspace{4pt}
\captionof{table}{\textbf{How is CAPE-JEPA different?}
\textbf{PP}: probabilistic prediction;
\textbf{DA}: distributional alignment;
\textbf{DG}: discriminative grounding;
\textbf{LS}: latent-space SSL;
\textbf{CM}: cross-modal RS image retrieval.}
\label{tab:comparison}
\vspace{-14pt}
\end{wrapfigure}

Joint Embedding Predictive Architectures
(JEPAs)~\cite{lecun2022path,assran2023self} offer a principled alternative: a
predictor $P_\phi$ estimates the target encoder's latent representation
$\mathbf{z}_t$ of masked patches from visible context $\mathbf{z}_c$, operating
entirely in semantic space without pixel decoding. I-JEPA~\cite{assran2023self}
established this paradigm for unimodal visual learning, and
REJEPA~\cite{choudhury2025rejepa} first adapted it to RS for unimodal retrieval.
X-JEPA~\cite{Choudhury_2026_WACV} extended the paradigm to cross-modal RS via a
cross-attentional predictor, shared learnable queries, and a geometry-aware
Prediction Space Alignment (PSA) loss. Despite strong results, X-JEPA exhibits
two fundamental limitations that motivate our work, summarised alongside related
methods in Table~\ref{tab:comparison} and through the theoretical analysis in Section~\ref{sec:method:theory}.

\emph{(1) Deterministic prediction under inherent cross-sensor uncertainty.}
X-JEPA's predictor regresses a single point estimate $\hat{\mathbf{z}}_t =
P_\phi(\mathbf{z}_c^{\text{local}}, \mathbf{z}_c^{\text{opp}})$ per masked
patch, where $\mathbf{z}_c^{\text{local}}$ and $\mathbf{z}_c^{\text{opp}}$
denote same-modality and opposite-modality context embeddings respectively,
implicitly treating all regions as equally predictable. In practice, clouds,
SAR speckle, mixed land-cover transitions, and acquisition noise introduce
substantial spatially non-uniform ambiguity: multiple valid latent
representations may correspond to a single masked patch depending on sensor
geometry and noise realisation. Forcing deterministic regression under such
ambiguity produces overconfident embeddings that conflate uncertain regions with
high-confidence predictions~\cite{kendall2017uncertainties,sluijterman2023optimal},
distorting the nearest-neighbour geometry that retrieval relies on. Following
Kendall and Gal~\cite{kendall2017uncertainties}, we distinguish \emph{aleatoric}
uncertainty (irreducible sensor noise, \eg SAR speckle) from \emph{epistemic}
uncertainty (model limitations given finite context), neither of which is representable by a point-estimate predictor.

\emph{(2) Geometrically disjoint modality distributions.}
X-JEPA's PSA loss enforces structural consistency between paired local
predictions but does not constrain the marginal embedding distributions
$P(\mathbf{z}^{(a)})$ and $P(\mathbf{z}^{(b)})$ globally, where superscripts
$(a)$ and $(b)$ index the two sensor modalities. Even when local pairwise geometry is preserved, the two modalities may occupy disjoint regions of the embedding manifold~\cite{liang2022mind}. This modality gap directly degrades nearest-neighbour retrieval, as queries and gallery items from different modalities are 
never co-located in the shared space.

\noindent\textbf{Research questions.}
These limitations motivate three concrete questions to address:
\begin{enumerate}[noitemsep, topsep=2pt]
    \item \emph{Can cross-sensor prediction uncertainty be explicitly modelled
    in latent space, and does doing so improve retrieval geometry?}
    \item \emph{Can the modality gap be closed globally without explicit sample
    pairing, and what is the right distributional objective to achieve this?}
    \item \emph{Are probabilistic prediction and distributional alignment
    individually sufficient, or must they be jointly trained with a
    discriminative objective to prevent collapse?}
\end{enumerate}

\noindent\textbf{Proposed approach.}
We propose \textbf{CAPE-JEPA} (\textbf{C}ross-modal \textbf{A}dversarial
\textbf{P}robabilistic Joint Embedding Predictive Architecture), which answers
all three questions through a unified probabilistic-distributional framework.
Rather than suppressing cross-sensor uncertainty, CAPE-JEPA replaces deterministic
latent regression with a Gaussian conditional predictor that outputs a per-patch
mean and variance for each masked region, trained via negative log-likelihood to
adaptively down-weight ambiguous patches while preserving strong supervision on
predictable regions. CAPE-JEPA then enforces global distributional convergence
between the two modality embedding distributions via adversarial characteristic
function tests~\cite{epps1983test,balestriero2025lejepa} that actively seek out
and penalise worst-case directions of cross-modal divergence, without requiring
explicit sample pairing. Finally, an in-training contrastive
objective~\cite{chen2020simple} with attention-pooled heads anchors scene-level
semantic structure within the geometrically aligned space, preventing
distributional regularization from collapsing inter-class discriminability.
In summary, our \textbf{major contributions} are:

\begin{itemize}[noitemsep, topsep=0pt]
    \item \textbf{Probabilistic cross-modal predictor:} The first
    uncertainty-aware latent predictor for RS-CMIR, modelling per-patch mean
    and variance via Gaussian NLL~\cite{kendall2017uncertainties,
    sluijterman2023optimal} to capture both aleatoric and epistemic
    cross-sensor uncertainty.

    \item \textbf{Adversarial distributional regularization:} A novel
    cross-modal regularizer (XSIGReg) that closes the modality
    gap~\cite{liang2022mind} via worst-case Epps-Pulley
    projections~\cite{epps1983test}, provably more powerful than random-projection
    approaches such as SIGReg~\cite{balestriero2025lejepa} (Standard Isotropic
    Gaussian Regularization) in the population limit.

    \item \textbf{Unified predictive-contrastive pretraining:} In-training
    InfoNCE~\cite{chen2020simple} with attention-pooled heads anchors
    scene-level discrimination within the aligned space, necessary to
    prevent semantic collapse.

    \item \textbf{Theoretical analysis and State-of-the-art on two RS-CMIR benchmarks:} Besides showing theoretical superiority against X-JEPA in Section~\ref{sec:method:theory}, CAPE-JEPA
    surpasses X-JEPA~\cite{Choudhury_2026_WACV} by up to \textbf{13.74} and
    \textbf{15.59 F1@5 points} on BEN-14K and fMoW, outperforming all
    baselines with fewer parameters
    (407.30M vs.\ 432.72M).
\end{itemize}
\vspace{-10pt}
\section{Related Work}
\label{sec:related}

\noindent\textbf{Self-Supervised Learning (SSL) for Remote Sensing.}
Contrastive methods~\cite{chen2020simple,he2020momentum} learn representations
by pulling together augmented views while pushing apart negatives, but are
fragile in RS: seasonal variation and sensor differences create false negatives
that destabilise training~\cite{Choudhury_2026_WACV,manas2021seasonal}. MIM
approaches~\cite{he2022masked,cong2022satmae,noman2024rethinking,hackstein2024exploring}
bypass this via masked pixel reconstruction, but pixel-space objectives
emphasise sensor-specific texture cues such as speckle, spectral profiles, resolution
artefacts, limiting cross-modal transferability~\cite{Choudhury_2026_WACV}.

\noindent\textbf{Cross-Modal Representation Learning.}
CROMA~\cite{fuller2023croma} combines contrastive and masked reconstruction
losses for SAR--multispectral alignment; SkySense~\cite{guo2024skysense} scales
this to multiple sensors via multi-stage pretraining; DeCUR~\cite{wang2024decoupling}
explicitly decouples shared semantic from modality-specific features via
redundancy-aware decorrelation. Despite strong alignment, these methods either
anchor learning to low-level pixel reconstruction or suffer from false-negative
sensitivity~\cite{Choudhury_2026_WACV}. CAPE-JEPA instead operates entirely in
latent space with probabilistic prediction and adversarial distributional
regularization.

\noindent\textbf{Remote Sensing Foundation Models.}
AnySat~\cite{astruc2025anysat}, Galileo~\cite{tseng2025galileo}, and
Panopticon~\cite{waldmann2025panopticon} demonstrate strong transfer across
diverse RS tasks via large-scale multi-sensor pretraining. However, all employ
deterministic prediction objectives and do not address cross-modal retrieval or
cross-sensor prediction uncertainty, the two gaps CAPE-JEPA specifically
targets.

\noindent\textbf{JEPA-Based Predictive Architectures.}
JEPAs~\cite{lecun2022path,assran2023self} reformulate SSL as latent-space
forecasting, yielding more abstract representations than pixel
reconstruction~\cite{assran2023self,bardes2024revisiting}. REJEPA~\cite{choudhury2025rejepa}
first adapted this to RS for unimodal retrieval; X-JEPA~\cite{Choudhury_2026_WACV}
extended it to the cross-modal setting via cross-attentional prediction and a
geometry-aware PSA loss. However, X-JEPA employs deterministic prediction and
leaves global cross-modal distributional alignment
unaddressed~\cite{liang2022mind}. CAPE-JEPA resolves both via probabilistic
Gaussian prediction~\cite{kendall2017uncertainties,sluijterman2023optimal},
adversarial XSIGReg~\cite{epps1983test}, and in-training
InfoNCE~\cite{chen2020simple}.

\noindent\textbf{Probabilistic and Uncertainty-Aware SSL.}
Kendall and Gal~\cite{kendall2017uncertainties} distinguish aleatoric from epistemic uncertainty, both relevant in noisy RS observations; Sluijterman et al.~\cite{sluijterman2023optimal} establish optimal NLL training
for mean-variance networks; probabilistic embeddings in supervised metric learning~\cite{shi2019probabilistic} show distributions better capture label ambiguity than point estimates. Yet uncertainty-aware learning remains
underexplored in SSL, where deterministic embeddings implicitly assume uniform patch predictability~\cite{assran2023self,Choudhury_2026_WACV}, an assumption particularly harmful in RS given clouds, SAR speckle, and mixed land-cover boundaries. CAPE-JEPA addresses this with probabilistic Gaussian prediction
trained via NLL.

\noindent\textbf{Distributional Alignment and Embedding Geometry.}
Uniformity and alignment objectives~\cite{wang2020understanding} prevent
collapse on the unit hypersphere; the Epps-Pulley
statistic~\cite{epps1983test} offers a non-parametric alternative via empirical
characteristic functions, applied in SIGReg~\cite{balestriero2025lejepa} to
enforce isotropic Gaussian structure on unimodal embeddings. CAPE-JEPA extends this to the cross-modal setting via XSIGReg, an adversarial two-sample EP test that targets worst-case projection directions of cross-modal divergence, which 
is more powerful than random projections and requiring no explicit sample pairing.



\section{Proposed Methodology}
\label{sec:method}

\subsection{Problem Formulation}
\label{sec:method:formulation}

Let $\mathcal{X} = \{x_i^{(a)}\}_{i=1}^{N}$ and
$\mathcal{Y} = \{x_i^{(b)}\}_{i=1}^{N}$ denote images from heterogeneous RS
modalities $a$ and $b$ (\eg, S1 SAR and S2 multispectral), where each pair
$(x_i^{(a)}, x_i^{(b)})$ captures semantically related scenes from different
sensors. The objective of RS-CMIR is to retrieve the most semantically relevant
images from modality $b$ given a query from modality $a$, and vice versa,
without human annotations or pixel-level correspondence.

Following the JEPA paradigm~\cite{lecun2022path,assran2023self}, we formulate
representation learning as a latent prediction task: given visible context
patches from both modalities, predict the latent embeddings of masked patches
conditioned on cross-modal context. Critically, unlike deterministic JEPA
formulations~\cite{assran2023self,Choudhury_2026_WACV} that regress a single
point estimate $\hat{\mathbf{z}}_t = P_\phi(\mathbf{z}_c)$, CAPE-JEPA models
this as a \emph{conditional Gaussian distribution}:
\begin{equation}
    p(\mathbf{z}_t \mid \mathbf{z}_c^{\text{local}}, \mathbf{z}_c^{\text{opp}})
    = \mathcal{N}\!\left(\boldsymbol{\mu}_\phi,\,
    \text{diag}(\boldsymbol{\sigma}^2_\phi)\right),
    \label{eq:conditional_gaussian}
\end{equation}
where $\boldsymbol{\mu}_\phi \in \mathbb{R}^{N_{\text{mask}} \times D}$ and
$\boldsymbol{\sigma}^2_\phi \in \mathbb{R}^{N_{\text{mask}} \times D}$ are the
predicted per-patch mean and variance, respectively, output by a
cross-attentional network conditioned on same-modality context
$\mathbf{z}_c^{\text{local}}$ and opposite-modality context
$\mathbf{z}_c^{\text{opp}}$. This formulation is the foundation of
CAPE-JEPA's superiority over X-JEPA: by explicitly representing prediction
uncertainty, the model avoids the overconfident embedding distortions that
arise when a deterministic predictor is forced to regress under inherent
cross-sensor ambiguity.

\begin{figure*}[t]
    \centering
    \setlength{\abovecaptionskip}{4pt}
    \setlength{\belowcaptionskip}{0pt}
    \includegraphics[width=\linewidth]{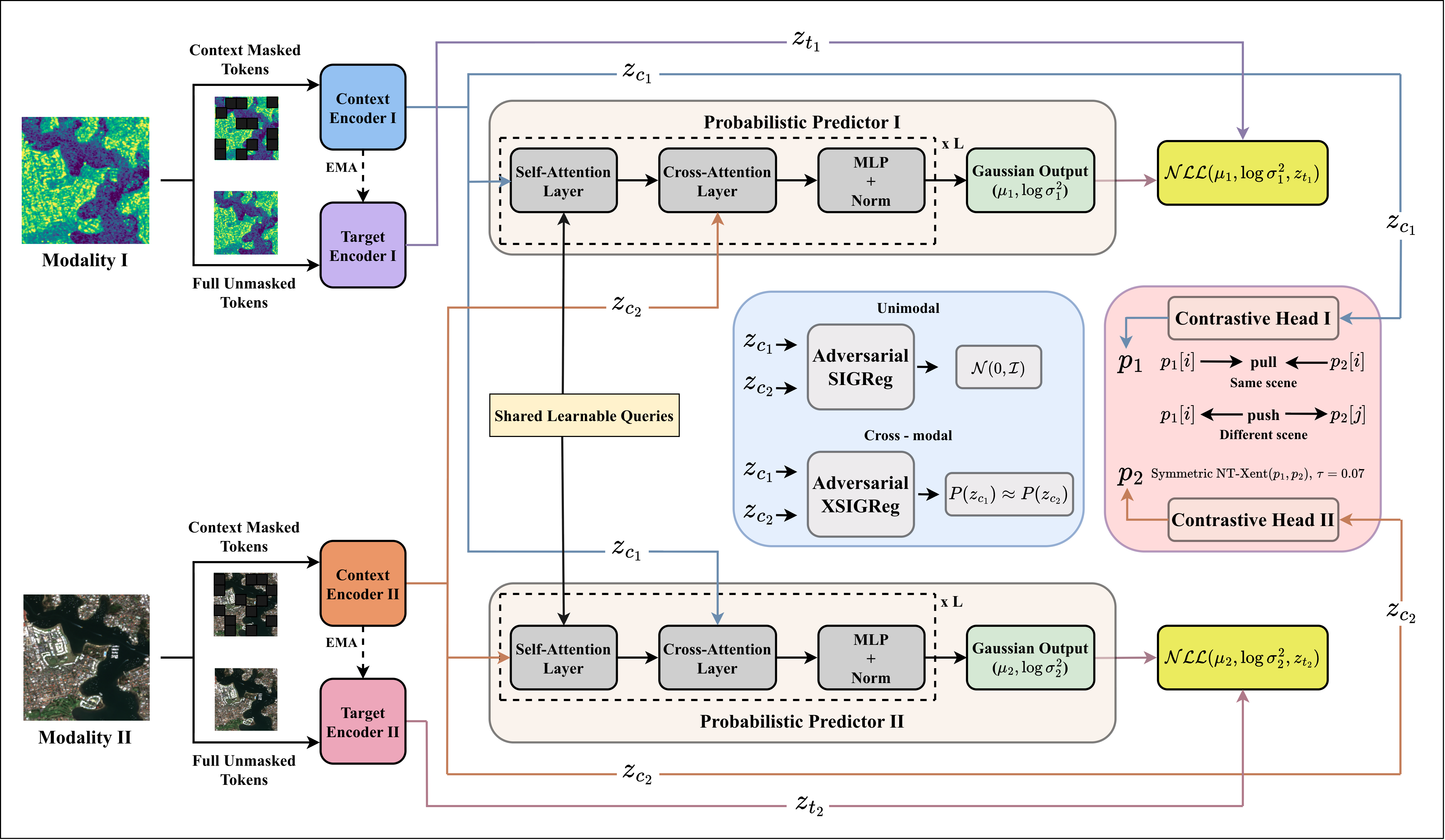}
    \caption{\textbf{Overview of CAPE-JEPA.} For each modality,
    a context encoder processes visible patches while an EMA target encoder
    generates stable prediction targets. The probabilistic cross-modal predictor
    outputs per-patch Gaussian parameters $(\boldsymbol{\mu},\boldsymbol{\sigma}^2)$
    conditioned on opposite-modality context. SIGReg enforces isotropic Gaussian
    structure on each modality's embeddings while XSIGReg aligns the cross-modal
    distributions via adversarial worst-case projections. In-training InfoNCE
    provides scene-level discriminative grounding.}
    \label{fig:architecture}
\end{figure*}

\vspace{-10pt}
\subsection{Architecture Overview}
\label{sec:method:arch}

CAPE-JEPA comprises four Vision Transformer (ViT)
encoders~\cite{dosovitskiy2020image} and two probabilistic cross-modal
predictors, as illustrated in Figure~\ref{fig:architecture}. For each modality
$s \in \{a, b\}$, a \textbf{context encoder} $E_\theta^{(s)}$ processes
visible patches, and an \textbf{EMA target encoder} $E_{\theta'}^{(s)}$
generates stable prediction targets~\cite{grill2020bootstrap,assran2023self}.
Target encoders are updated via exponential moving average:
\begin{equation}
    \theta' \leftarrow m\,\theta' + (1-m)\,\theta,
    \label{eq:ema}
\end{equation}
with momentum $m$ annealed from $0.996$ to $1.0$. Target encoder parameters
receive no gradients, ensuring stable regression targets throughout training.

\noindent\textbf{Patch Encoding.}
Each $H \times W$ image is divided into non-overlapping $16{\times}16$ patches,
yielding $N = HW/256$ tokens. Each patch is linearly projected to a
$D$-dimensional embedding and summed with a learnable sinusoidal positional
encoding~\cite{vaswani2017attention}, producing $\mathbf{x} \in
\mathbb{R}^{N \times D}$. A prepended CLS token aggregates global context but
is excluded from pooling at inference.

\noindent\textbf{Complementary Masking.}
We partition patches into context and target sets via a complementary strategy.
A binary mask $\mathcal{M}^{(a)} \subset \{1,\ldots,N\}$ is sampled for
modality $a$ with ratio $r \in [0.15, 0.25]$; modality $b$'s context mask is
its complement: $\mathcal{M}^{(b)} = \{1,\ldots,N\} \setminus \mathcal{M}^{(a)}$.
What modality $a$ observes as context, modality $b$ must predict, and vice
versa. Here $\mathbf{z}_c^{(b,\text{local})}$ denotes the local
same-modality context of modality $b$, i.e., the unmasked patches
of modality $b$. Unlike the independent masking of X-JEPA~\cite{Choudhury_2026_WACV}, this hard structural constraint forces each 
predictor to genuinely exploit opposite-modality features rather than 
reconstructing from local context alone.
We note that the ablation in Table~\ref{tab:ablation_masking} shows complementary masking
underperforms on S2$\to$S1 in isolation; this reflects that independent masking
preserves more local S2 context which benefits SAR retrieval when S2 features
are already discriminative, but complementary masking yields superior
cross-modal performance overall with the full objective. Theoretical justification for complementary masking is
provided in \textbf{Section~5} of the supplementary.

\noindent\textbf{Inference.}
The context encoder processes the full unmasked image. Patch tokens (excluding
CLS) are mean-pooled to a single $D$-dimensional embedding, optionally
projected through the alignment head, and compared via cosine similarity for
nearest-neighbour retrieval.

\vspace{-10pt}
\subsection{Probabilistic Cross-Modal Predictor}
\label{sec:method:predictor}

The core architectural novelty of CAPE-JEPA is replacing the deterministic
predictor of X-JEPA with a \textbf{probabilistic Gaussian
predictor}~\cite{kendall2017uncertainties,sluijterman2023optimal}. The
motivation is principled: in RS imagery, clouds, SAR speckle, mixed land-cover
boundaries, and acquisition noise mean that a single masked patch $x_t^{(a)}$
has multiple plausible latent representations given any finite opposite-modality
context. Forcing a point estimate onto this one-to-many mapping produces
systematically overconfident embeddings that distort nearest-neighbour geometry.
Modelling the full conditional distribution (Eq.~\ref{eq:conditional_gaussian})
resolves this by adaptively allocating predictive confidence in proportion to
regional ambiguity.

\noindent\textbf{CrossAttentionBlock.}
Each block performs three sequential
operations~\cite{vaswani2017attention}: self-attention over query tokens,
cross-attention where query tokens attend to opposite-modality context
$\mathbf{c}^{\text{opp}}$ as keys and values, and an MLP with layer
normalisation. In the cross-attention step, the query $\mathbf{x}$ is not separately layer-normalised as it has already passed through the self-attention residual which applies layer normalisation; only the opposite-modality context $\mathbf{c}^{\text{opp}}$ is normalised before serving as keys and values.

\noindent\textbf{Shared Learnable Queries.}
Each predictor maintains $Q{=}8$ learnable query tokens $\mathbf{s} \in
\mathbb{R}^{Q \times P}$ acting as modality-agnostic semantic anchors, enabling
consistent cross-modal information routing~\cite{Choudhury_2026_WACV}.

\noindent\textbf{Token Sequence and Gaussian Output.}
The predictor input concatenates shared queries, projected local context tokens,
and positionally-encoded mask tokens:
\begin{equation}
    \mathbf{q} = \bigl[\,\mathbf{s}\;\|\;\text{proj}(\mathbf{z}_c^{\text{local}})
    \;\|\;(\mathbf{e}_{\text{mask}} + \text{proj}(\mathbf{e}_{\text{pos}}))\,\bigr],
    \label{eq:tokens}
\end{equation}

where $\mathbf{z}_c^{\text{local}} \in \mathbb{R}^{(N - N_{\text{mask}})
\times D}$ are context tokens from the query modality, $\mathbf{e}_{\text{mask}} \in \mathbb{R}^{N_{\text{mask}} \times D}$ is a learnable mask
token~\cite{he2022masked}, and $\mathbf{e}_{\text{pos}} \in \mathbb{R}^{N
\times D}$ encodes patch positions. Opposite-modality context
$\mathbf{z}_c^{\text{opp}}$ enters as cross-attention keys and values in each
block. After $L{=}12$ stacked blocks, the final $N_{\text{mask}}$ tokens are
projected to $2D$ dimensions and split:
\begin{equation}
    \bigl(\boldsymbol{\mu}_\phi,\,\log\boldsymbol{\sigma}^2_\phi\bigr)
    = P_\phi\!\left(\mathbf{z}_c^{\text{local}},\,\mathbf{z}_c^{\text{opp}}\right),
    \label{eq:predictor_out}
\end{equation}
with $\boldsymbol{\mu}_\phi,\,\log\boldsymbol{\sigma}^2_\phi \in
\mathbb{R}^{B \times N_{\text{mask}} \times D}$, where $B$ is the batch size.

\noindent\textbf{Gaussian NLL Loss.}
The predictor maximises the likelihood of target embeddings $\mathbf{z}_t =
E_{\theta'}(\mathbf{x}_t)$ under the predicted
distribution~\cite{sluijterman2023optimal}:
\begin{equation}
    \mathcal{L}_{\text{NLL}} = \mathbb{E}\!\left[
    \frac{(\mathbf{z}_t - \boldsymbol{\mu}_\phi)^2}{2\boldsymbol{\sigma}^2_\phi}
    + \frac{1}{2}\log\boldsymbol{\sigma}^2_\phi\right],
    \label{eq:nll}
\end{equation}
where the expectation is over samples and masked patches, and operations are
element-wise over latent dimensions $d \in \{1,\ldots,D\}$. The explicit
averaging constant $\frac{1}{B \cdot N_{\text{mask}} \cdot D}$ is given
in the \textbf{supplementary material}. The $\log
\boldsymbol{\sigma}^2_\phi$ term prevents variance collapse to zero, while
large $\boldsymbol{\sigma}^2_\phi$ incurs a regularization penalty that
discourages uninformative predictions. These two competing forces adaptively
down-weight ambiguous patches without discarding their supervision
signal, yielding a strictly richer learning objective than X-JEPA's L2 regression,
as formalised in Proposition~1 (Section~\ref{sec:method:theory}).

\vspace{-10pt}
\subsection{Distributional Regularization}
\label{sec:method:sigreg}

$\mathcal{L}_{\text{NLL}}$ governs patch-level uncertainty but leaves the
\emph{global} embedding geometry unconstrained~\cite{liang2022mind}. Collapsed
dimensions, anisotropy, or modality-specific offset (the modality gap) distort
nearest-neighbour relations even when local predictions are
accurate~\cite{wang2020understanding}. We address this with two complementary
regularizers at different levels of granularity.

\noindent\textbf{Unimodal SIGReg.}
We adopt SIGReg~\cite{balestriero2025lejepa} to enforce approximately isotropic
Gaussian structure on each modality's context embeddings $\mathbf{z}_c^{(s)}
\in \mathbb{R}^D$ independently. Random one-dimensional projections $y_i =
\mathbf{z}_i^{(s)\top}\mathbf{v}$, $\mathbf{v} \sim \mathcal{N}(\mathbf{0},
\mathbf{I})$, are compared to a standard Gaussian via the Epps-Pulley (EP)
characteristic function statistic $\mathcal{D}_{\text{EP}}$~\cite{epps1983test},
which measures distributional discrepancy non-parametrically via the empirical
characteristic function. This is averaged over $K{=}64$ random directions:
\begin{equation}
    \mathcal{L}_{\text{SIG}} = \frac{1}{K}\sum_{k=1}^{K}
    \mathcal{D}_{\text{EP}}\!\left(\bigl\{\mathbf{z}_i^{(s)\top}
    \mathbf{v}_k\bigr\}_{i=1}^{B}\right),\quad s \in \{a,b\}.
    \label{eq:sigreg}
\end{equation}

\noindent\textbf{Cross-Modal XSIGReg.}
SIGReg regularizes each modality independently but does not close the gap
between $P(\mathbf{z}_c^{(a)})$ and $P(\mathbf{z}_c^{(b)})$. Moreover,
random projections may systematically miss the geometrically significant
directions along which the two distributions diverge most in high dimensions.
We address both limitations with a novel \textbf{adversarial two-sample EP
test}~\cite{epps1983test,goodfellow2014generative}: an adversarial projection
direction $\mathbf{v}^* \in \mathbb{S}^{D-1}$ is sought to maximise cross-modal
distributional discrepancy, while encoder parameters minimise it:
\begin{equation}
    \mathcal{L}_{\text{XSIG}} = \max_{\mathbf{v} \in \mathbb{S}^{D-1}}
    \mathcal{D}_{\text{EP}}\!\left(
    \bigl\{\mathbf{z}_i^{(a)\top}\mathbf{v}\bigr\},\,
    \bigl\{\mathbf{z}_i^{(b)\top}\mathbf{v}\bigr\}
    \right).
    \label{eq:xsigreg}
\end{equation}
This min-max saddle-point problem~\cite{goodfellow2014generative} is optimised
by alternating gradient steps: $\mathbf{v}$ is updated by gradient ascent
with step size $\eta_{\text{adv}}{=}0.01$ and re-projected onto $\mathbb{S}^{D-1}$ after
each step; encoder parameters are updated by gradient descent. $\mathbf{v}$ is
re-initialised from $\mathcal{N}(\mathbf{0}, \mathbf{I})$ at the start of each
training iteration. We perform $T{=}5$ ascent steps per encoder update, sufficient for
stable convergence in practice; the EP statistic's smoothness and
boundedness~\cite{epps1983test} provide more favourable optimisation
landscape properties than unbounded GAN-style
objectives~\cite{goodfellow2014generative}. Unlike SIGReg's random projections, XSIGReg actively identifies and
closes the worst-case cross-modal divergence direction, making it suitable for eliminating the modality gap~\cite{liang2022mind}.

\subsection{Scene-Level Discriminative Supervision}
\label{sec:method:nce}

Distributional regularization aligns the geometry of the embedding space but
provides no between-class discriminative signal: both modalities may converge
to the same well conditioned distribution while remaining semantically
flat~\cite{liang2022mind,chen2020simple}. We resolve this tension with
\textbf{in-training InfoNCE}~\cite{chen2020simple} computed directly during
pretraining.

\noindent\textbf{Contrastive Head.}
Each modality has a dedicated contrastive head: a learned-query attention
pooling layer~\cite{lee2019set} followed by a two-layer MLP $f_s :
\mathbb{R}^D \rightarrow \mathbb{R}^{512} \rightarrow \mathbb{R}^{256}$ with
GELU as the activation function to maintain consistency with the
ViT backbone~\cite{dosovitskiy2020image} and preserve smooth gradient
flow for the probabilistic representations.
Let $\mathbf{h}_i^{(s)} = f_s\!\left(\text{AttnPool}(\mathbf{z}_{c,i}^{(s)})\right)$
denote the unprojected representation of scene $i$ from modality $s$. The
$\ell_2$-normalised projection is then:
\begin{equation}
    \mathbf{p}_i^{(s)} = \frac{\mathbf{h}_i^{(s)}}
    {\left\|\mathbf{h}_i^{(s)}\right\|_2},
    \quad s \in \{a,b\}.
    \label{eq:contrastive_head}
\end{equation}

\noindent\textbf{Symmetric NT-Xent Loss.}
Given paired projections $\{\mathbf{p}_i^{(a)}, \mathbf{p}_i^{(b)}\}_{i=1}^B$,
where $\mathbf{p}_i^{(s)} \in \mathbb{R}^{256}$ denotes the $\ell_2$-normalised
projection of scene $i$ from modality $s$:
\begin{equation}
    \mathcal{L}_{\text{NCE}} = -\frac{1}{2B}\sum_{i=1}^{B}\!\left[
    \log\frac{e^{\mathbf{p}_i^{(a)\top}\mathbf{p}_i^{(b)}/\tau}}
    {\sum_{j=1}^{B}e^{\mathbf{p}_i^{(a)\top}\mathbf{p}_j^{(b)}/\tau}}
    +
    \log\frac{e^{\mathbf{p}_i^{(b)\top}\mathbf{p}_i^{(a)}/\tau}}
    {\sum_{j=1}^{B}e^{\mathbf{p}_i^{(b)\top}\mathbf{p}_j^{(a)}/\tau}}
    \right],
    \label{eq:nce}
\end{equation}
with temperature $\tau{=}0.07$~\cite{chen2020simple}. Positives are paired
cross-modal embeddings of the same scene; negatives are all other $B{-}1$
in-batch samples. Both directions (modality $a$ querying $b$, and vice versa)
are computed symmetrically.

\vspace{-10pt}
\subsection{Total Training Objective}
\label{sec:method:loss}

The complete CAPE-JEPA objective combines all four terms:
\begin{equation}
    \mathcal{L} = \mathcal{L}_{\text{NLL}} + \lambda_1\,\mathcal{L}_{\text{SIG}}
    + \lambda_2\,\mathcal{L}_{\text{XSIG}} + \lambda_3\,\mathcal{L}_{\text{NCE}},
    \label{eq:total_loss}
\end{equation}
with the $\lambda$s selected through grid search.
$\mathcal{L}_{\text{NLL}}$ and $\mathcal{L}_{\text{SIG}}$ are applied
symmetrically to both modalities. Only context encoders and predictors receive
gradients; target encoders are updated solely via EMA (Eq.~\ref{eq:ema}).
The three objectives operate at complementary levels of granularity:
$\mathcal{L}_{\text{NLL}}$ at the patch level,
$\mathcal{L}_{\text{SIG}}$/$\mathcal{L}_{\text{XSIG}}$ at the distribution
level, and $\mathcal{L}_{\text{NCE}}$ at the scene level. As confirmed by the
component ablation (Table~\ref{tab:ablation_components}), each term is
load-bearing: XSIGReg without InfoNCE causes cross-modal performance degradation
($-2.03$ on S2$\to$S1) because distributional alignment without discriminative
grounding erases semantic separation; InfoNCE without XSIGReg collapses
cross-modal geometry ($-8.89$ on S2$\to$S1) because contrastive clusters resist
distributional alignment. Only the full objective satisfies all
three levels.

\vspace{-10pt}
\subsection{Post-Pretraining Alignment}
\label{sec:method:align}

Following pretraining, context encoders are frozen and contrastive heads are
fine-tuned using $\mathcal{L}_{\text{NCE}}$ (Eq.~\ref{eq:nce}) with paired
cross-modal samples as positives. This alignment stage is an integral part of the CAPE-JEPA training pipeline and is not applicable to baselines. Notably, even without
this alignment stage (0 epochs, Table~\ref{tab:ablation_align}),
CAPE-JEPA achieves S1$\to$S2 = 73.57 and S2$\to$S1 = 66.02
vs X-JEPA's 62.74 and 62.98 respectively, confirming
that the core pretraining objective itself drives the improvement
independently of the alignment stage. 
\subsection{Theoretical Analysis: Why CAPE-JEPA Outperforms X-JEPA}
\label{sec:method:theory}

We formalise the advantage of CAPE-JEPA over X-JEPA through three arguments.

\noindent\textbf{Proposition 1 (NLL subsumes L2 under uncertainty).}
Let $\mathbf{z}_t$ be a target embedding and let the predictor output
$(\boldsymbol{\mu}_\phi, \boldsymbol{\sigma}^2_\phi)$. The Gaussian NLL
(Eq.~\ref{eq:nll}) reduces to scaled L2 when
$\boldsymbol{\sigma}^2_\phi = c\,\mathbf{1}$ for constant $c > 0$, recovering
X-JEPA's objective as a special case. Under non-uniform
predictability, which is the realistic RS setting, minimising
$\mathcal{L}_{\text{NLL}}$ yields lower expected retrieval distortion
than L2, since it adaptively re-weights each patch's gradient by
$1/\boldsymbol{\sigma}^2_\phi$, suppressing the influence of inherently
ambiguous patches. We note this constitutes a design motivation
rather than a formal bound.

\noindent\textbf{Proposition 2 (XSIGReg closes the modality gap).}
Let $\Delta_{\text{gap}}$ (Eq.~\ref{eq:delta_gap}) denote the worst-case
projected distributional discrepancy between modalities:
\begin{equation}
    \Delta_{\text{gap}} = \sup_{\mathbf{v} \in \mathbb{S}^{D-1}}
    \mathcal{D}_{\text{EP}}\!\left(
    \bigl\{\mathbf{z}_i^{(a)\top}\mathbf{v}\bigr\},\,
    \bigl\{\mathbf{z}_i^{(b)\top}\mathbf{v}\bigr\}
    \right).
    \label{eq:delta_gap}
\end{equation}
X-JEPA's unbounded PSA loss does not penalise $\Delta_{\text{gap}}$. $\mathcal{L}_{\text{XSIG}}$ directly minimises $\Delta_{\text{gap}}$
via the saddle-point update, driving both modality distributions toward the same
isotropic Gaussian, which is a necessity for unbiased nearest-neighbour
retrieval across modalities \cite{liang2022mind}. Empirically, CAPE-JEPA
achieves a significantly lower Fr\'{e}chet distance between S1 and S2
distributions (CAPE-JEPA: \textbf{10.8} vs.\ X-JEPA: \textbf{22.9}),
confirming that XSIGReg closes the modality gap in practice
(see \textbf{Section~6} of the supplementary).

\noindent\textbf{Proposition 3 (Complementary necessity of three-level
objectives).}
We decompose the retrieval error as $\epsilon = \epsilon_{\text{patch}} +
\epsilon_{\text{dist}} + \epsilon_{\text{disc}}$, where
$\epsilon_{\text{patch}}$ measures patch-level prediction distortion (deviation
of $\boldsymbol{\mu}_\phi$ from $\mathbf{z}_t$ weighted by ambiguity),
$\epsilon_{\text{dist}}$ measures the modality gap
$\Delta_{\text{gap}}$ (Eq.~\ref{eq:delta_gap}), and $\epsilon_{\text{disc}}$
measures between-class discriminative failure (overlap of class-conditional
embedding distributions). $\mathcal{L}_{\text{NLL}}$ bounds
$\epsilon_{\text{patch}}$ but leaves $\epsilon_{\text{dist}}$ and
$\epsilon_{\text{disc}}$ unconstrained; $\mathcal{L}_{\text{SIG}} +
\mathcal{L}_{\text{XSIG}}$ bound $\epsilon_{\text{dist}}$ but provide no
discriminative signal, allowing $\epsilon_{\text{disc}}$ to grow;
$\mathcal{L}_{\text{NCE}}$ bounds $\epsilon_{\text{disc}}$ but does not
constrain $\epsilon_{\text{dist}}$. Only the full objective
(Eq.~\ref{eq:total_loss}) simultaneously bounds all three components.
This decomposition is a design rationale grounded in the functional roles of
each loss term; the empirical validity of the argument is confirmed by
the component ablation (Table~\ref{tab:ablation_components}), where
removing any single term degrades precisely the error component it governs,
and this pattern holds consistently under both F1@5 and mAP
(\textbf{Table~1} of the supplementary). Extended theoretical analysis is provided in \textbf{Section~4} of the supplementary.

\begin{table*}[t]
\centering
\setlength{\abovecaptionskip}{4pt}
\setlength{\belowcaptionskip}{6pt}
\caption{Retrieval performance (F1@5, \%) on BEN-14K and fMoW. Best results
in \textbf{bold}. $\dagger$~denotes methods not originally designed for
cross-modal retrieval. Methods are grouped by paradigm:
\colorbox{maerow!80}{\strut MIM-based},
\colorbox{crossrow!80}{\strut cross-modal SSL}, and
\colorbox{jeparow!80}{\strut JEPA-based}. ($^\ddagger$see footnote)}
\label{tab:main}
\setlength{\tabcolsep}{5pt}
\renewcommand{\arraystretch}{1.25}
\resizebox{0.95\textwidth}{!}{%
\begin{tabular}{lc cccc cccc}
\toprule
\multirow{3}{*}{\textbf{Method}} &
\multirow{3}{*}{\textbf{Params (M)}} &
\multicolumn{4}{c}{\cellcolor{benblue}\textbf{BEN-14K}} &
\multicolumn{4}{c}{\cellcolor{fmowgreen}\textbf{fMoW}} \\
\cmidrule(lr){3-6}\cmidrule(lr){7-10}
& &
\multicolumn{2}{c}{\cellcolor{unimodal}\textit{Unimodal}} &
\multicolumn{2}{c}{\cellcolor{crossmodal}\textit{Cross-modal}} &
\multicolumn{2}{c}{\cellcolor{unimodal}\textit{Unimodal}} &
\multicolumn{2}{c}{\cellcolor{crossmodal}\textit{Cross-modal}} \\
\cmidrule(lr){3-4}\cmidrule(lr){5-6}\cmidrule(lr){7-8}\cmidrule(lr){9-10}
& &
\cellcolor{unimodal}S1$\to$S1 & \cellcolor{unimodal}S2$\to$S2 &
\cellcolor{crossmodal}S1$\to$S2 & \cellcolor{crossmodal}S2$\to$S1 &
\cellcolor{unimodal}RGB$\to$RGB & \cellcolor{unimodal}SEN$\to$SEN &
\cellcolor{crossmodal}RGB$\to$SEN & \cellcolor{crossmodal}SEN$\to$RGB \\
\midrule
\rowcolor{maerow!50}
MAE$^\dagger$~\cite{he2022masked}
  & 224.87 & 60.81 & 72.04 & 41.78 & 46.12 & 58.73 & 61.18 & 34.23 & 35.12 \\
\rowcolor{maerow!50}
MAE-RVSA$^\dagger$~\cite{wang2022advancing}
  & 227.75 & 55.40 & 71.47 & 36.66 & 38.05 & 55.26 & 60.28 & 40.46 & 38.65 \\
\rowcolor{maerow!50}
SatMAE$^\dagger$~\cite{cong2022satmae}
  & 329.40 & 70.86 & 78.71 & 49.57 & 52.48 & 61.68 & 56.63 & 50.21 & 51.05 \\
\rowcolor{maerow!50}
SatMAE++$^\dagger$~\cite{noman2024rethinking}
  & 329.14 & 67.29 & 76.48 & 50.21 & 54.98 & 60.09 & 57.75 & 50.39 & 50.03 \\
\rowcolor{maerow!50}
CrossMAE$^\dagger$~\cite{fu2024rethinking}
  & 250.57 & 66.45 & 71.28 & 49.46 & 48.71 & 66.71 & 63.46 & 48.21 & 51.78 \\
\rowcolor{maerow!50}
CSMAE-SESD~\cite{hackstein2024exploring}
  & 210.64 & 70.62 & 39.01 & 38.74 & 38.42 & 51.56 & 41.09 & 30.87 & 30.84 \\
\midrule
\rowcolor{crossrow!50}
SkySense~\cite{guo2024skysense}
  & 398.04 & 69.87 & 73.42 & 50.26 & 52.11 & 68.81 & 67.45 & 51.62 & 52.41 \\
\rowcolor{crossrow!50}
CROMA~\cite{fuller2023croma}
  & 310.54 & 68.48 & 72.71 & 46.53 & 48.61 & 65.23 & 66.73 & 49.71 & 47.68 \\
\rowcolor{crossrow!50}
DeCUR~\cite{wang2024decoupling}
  & 250.54 & 71.26 & 75.36 & 40.78 & 41.83 & 63.42 & 60.17 & 48.12 & 45.43 \\
\midrule
\rowcolor{jeparow!50}
REJEPA~\cite{choudhury2025rejepa}
  & 197.09 & 76.38 & 75.42 & 55.46 & 56.32 & 73.53 & 75.87 & 50.08 & 49.85 \\
\rowcolor{jeparow!50}
X-JEPA~\cite{Choudhury_2026_WACV}
  & 432.72$^\ddagger$ & 72.31 & 77.24 & 62.74 & 62.98 & 73.59 & 77.08 & 52.86 & 54.68 \\

\midrule

\rowcolor{ourrow!30} 
CAPE-JEPA (MSE) & 406.52 & 73.50 & 70.44 & 66.34 & 65.71 & 73.21 & 74.16 & 56.49 & 62.83 \\

\rowcolor{ourrow!80}
\textbf{CAPE-JEPA (ours)}
  & \textbf{407.30}
  & \textbf{79.22} & \textbf{80.64}
  & \textbf{76.48} & \textbf{76.50}
  & \textbf{77.76} & \textbf{79.06}
  & \textbf{61.02} & \textbf{70.27} \\
\bottomrule
\end{tabular}%
}
\begin{minipage}{0.95\textwidth}
\vspace{4pt}
\small $^\ddagger$X-JEPA~\cite{Choudhury_2026_WACV} originally reported trainable parameters only, excluding EMA target encoders and predictor; we report the full parameter count for fair comparison. CAPE-JEPA~(MSE) denotes the deterministic variant replacing $\mathcal{L}_{\text{NLL}}$ with MSE loss.
\end{minipage}
\end{table*}


\vspace{-10pt}
\section{Experimental Evaluations}
\label{sec:experiments}

\noindent \textbf{Datasets Description.} We validate CAPE-JEPA on the following benchmark datasets, following the literature \cite{Choudhury_2026_WACV}:
\textbf{(i) BigEarthNet-14K (BEN-14K).}
BEN-14K~\cite{sumbul2019bigearthnet} contains 14,832 strictly paired 
Sentinel-1 (S1) SAR~\cite{Torres2012TheSM} and Sentinel-2 (S2) 
multispectral~\cite{drusch2012sentinel} patches. S1 images provide two polarisation channels (VV: vertical-vertical, VH: vertical-horizontal) at 10\,m resolution; S2 images provide 10 
spectral bands at 10--20\,m resolution. Each patch is annotated with 19 
multi-label land-cover classes following the CORINE Land Cover 
taxonomy~\cite{Bossard2000CORINELC}. We use the standard train/val/test split 
and evaluate on the validation set following the protocol
of~\cite{Choudhury_2026_WACV}, where the validation set is used
for final reporting to ensure comparability with prior work.
Hyperparameters are tuned on the training set only.
\textbf{(ii) fMoW-Sentinel.}
fMoW-Sentinel~\cite{Christie_2018_CVPR,cong2022satmae} extends the Functional 
Map of the World dataset with Sentinel-2 multispectral 
imagery~\cite{drusch2012sentinel} (13 channels) paired with RGB optical images 
(3 channels) across 62 semantic categories. Unlike BEN-14K, fMoW uses 
class-mode pairing: images from the same class are treated as semantic positives 
during evaluation. The dataset contains approximately 363,000 training and 
53,000 test images. Following \cite{Choudhury_2026_WACV}, we evaluate on the validation set. Extended dataset descriptions and evaluation protocol details are provided in \textbf{Section~1} of the supplementary.

\noindent\textbf{Selection of Competitors.}
We compare against eleven self-supervised RS methods: MAE~\cite{he2022masked},
MAE-RVSA~\cite{wang2022advancing}, SatMAE~\cite{cong2022satmae},
SatMAE++~\cite{noman2024rethinking}, CrossMAE~\cite{fu2024rethinking},
CSMAE-SESD~\cite{hackstein2024exploring}, SkySense~\cite{guo2024skysense},
CROMA~\cite{fuller2023croma}, DeCUR~\cite{wang2024decoupling},
REJEPA~\cite{choudhury2025rejepa}, and X-JEPA~\cite{Choudhury_2026_WACV}.
All methods use ViT-B/16~\cite{dosovitskiy2020image} except
SkySense (398M, its own backbone). Results are reported as F1@5, standard
for RS-CMIR evaluation~\cite{Choudhury_2026_WACV}; mAP results confirming
consistent trends are provided in the supplementary material.

\noindent\textbf{Implementation Details.}
CAPE-JEPA uses a ViT-B backbone~\cite{dosovitskiy2020image} ($D{=}768$, depth
12, 12 heads) for all four encoders, with predictor hidden size $P{=}384$
(which also serves as the shared query embedding dimension in
Eq.~\ref{eq:tokens}), depth 12, and $Q{=}8$ shared learnable queries. Complementary masking ratio
is 0.15--0.25. We train for 300 epochs with
AdamW~\cite{loshchilov2017decoupled}, peak learning rate $1{\times}10^{-3}$,
batch size 64, cosine decay~\cite{loshchilov2016sgdr}, and 2-epoch warmup.
EMA momentum is annealed from 0.996 to
1.0~\cite{grill2020bootstrap,assran2023self}. Loss weights $\lambda_1{=}0.01$,
$\lambda_2{=}0.0005$, $\lambda_3{=}0.15$; sensitivity analysis is in the
supplementary. Projection heads are aligned post-pretraining using
InfoNCE~\cite{chen2020simple}. Experiments run on four NVIDIA A100
GPUs. Additional implementation details including variance clamping,
NLL averaging, and XSIGReg hyperparameters are provided in
\textbf{Section~3} of the supplementary.

\begin{figure*}[t]
\centering
\setlength{\abovecaptionskip}{4pt}
\setlength{\belowcaptionskip}{0pt}
\includegraphics[width=0.95\textwidth, keepaspectratio]{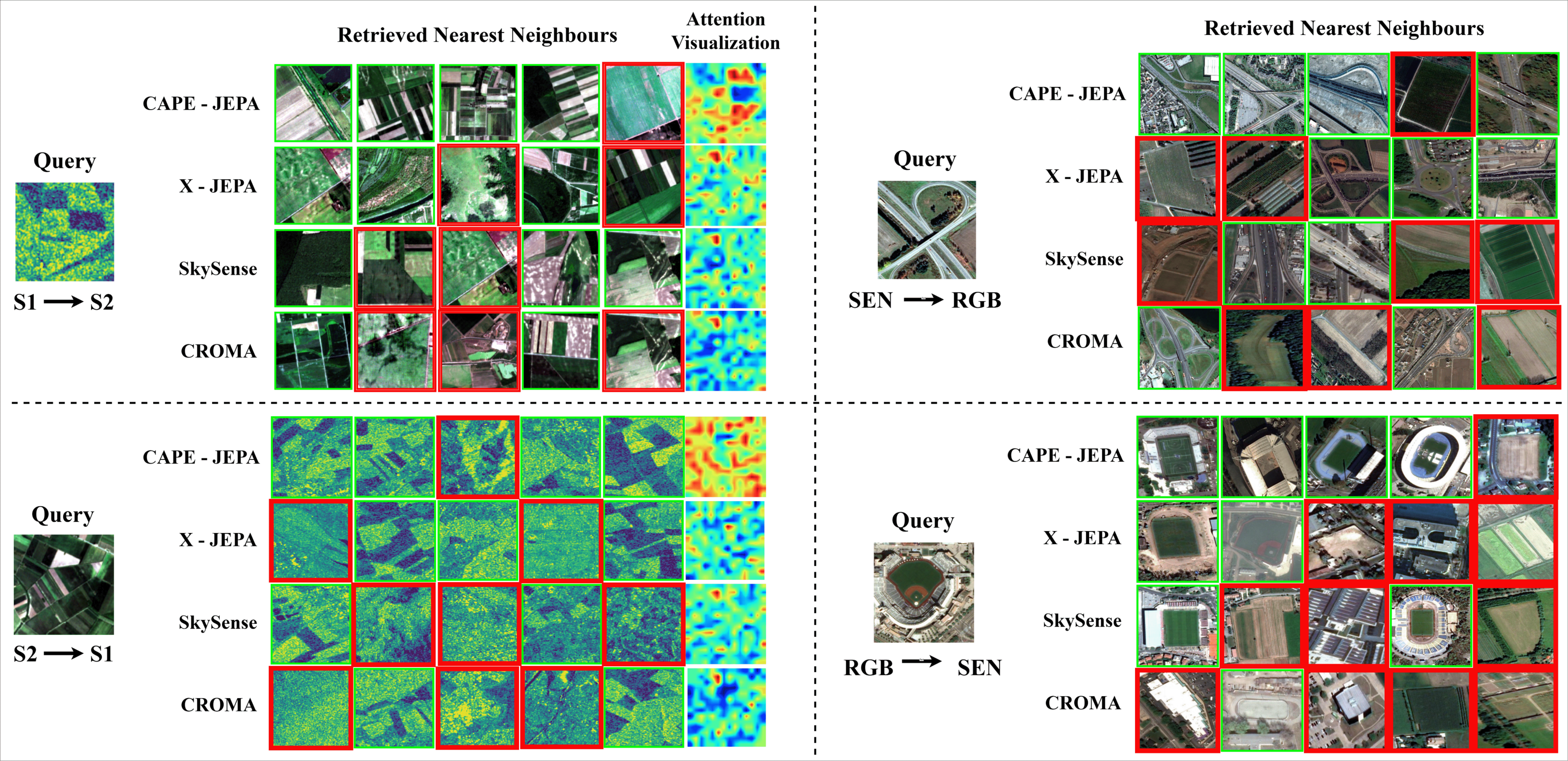}
\caption{\textbf{Qualitative cross-modal retrieval results} for queries on 
BEN-14K (S1$\rightarrow$S2 and S2$\rightarrow$S1, left) and fMoW 
(SEN$\rightarrow$RGB and RGB$\rightarrow$SEN, right). The query image 
is identical across all methods. Retrieved images are marked with 
\textcolor{green}{\textbf{green}} borders for correct retrievals and 
\textcolor{red}{\textbf{red}} borders for incorrect retrievals. The attention visualization column (BEN-14K) shows cross-modal attention  maps from the final cross-attention layer, highlighting discriminative regions.}
\label{fig:qualitative}
\end{figure*}

\begin{figure*}[t]
\centering
\setlength{\abovecaptionskip}{5pt}
\setlength{\belowcaptionskip}{1pt}
\includegraphics[width=0.75\textwidth, keepaspectratio]{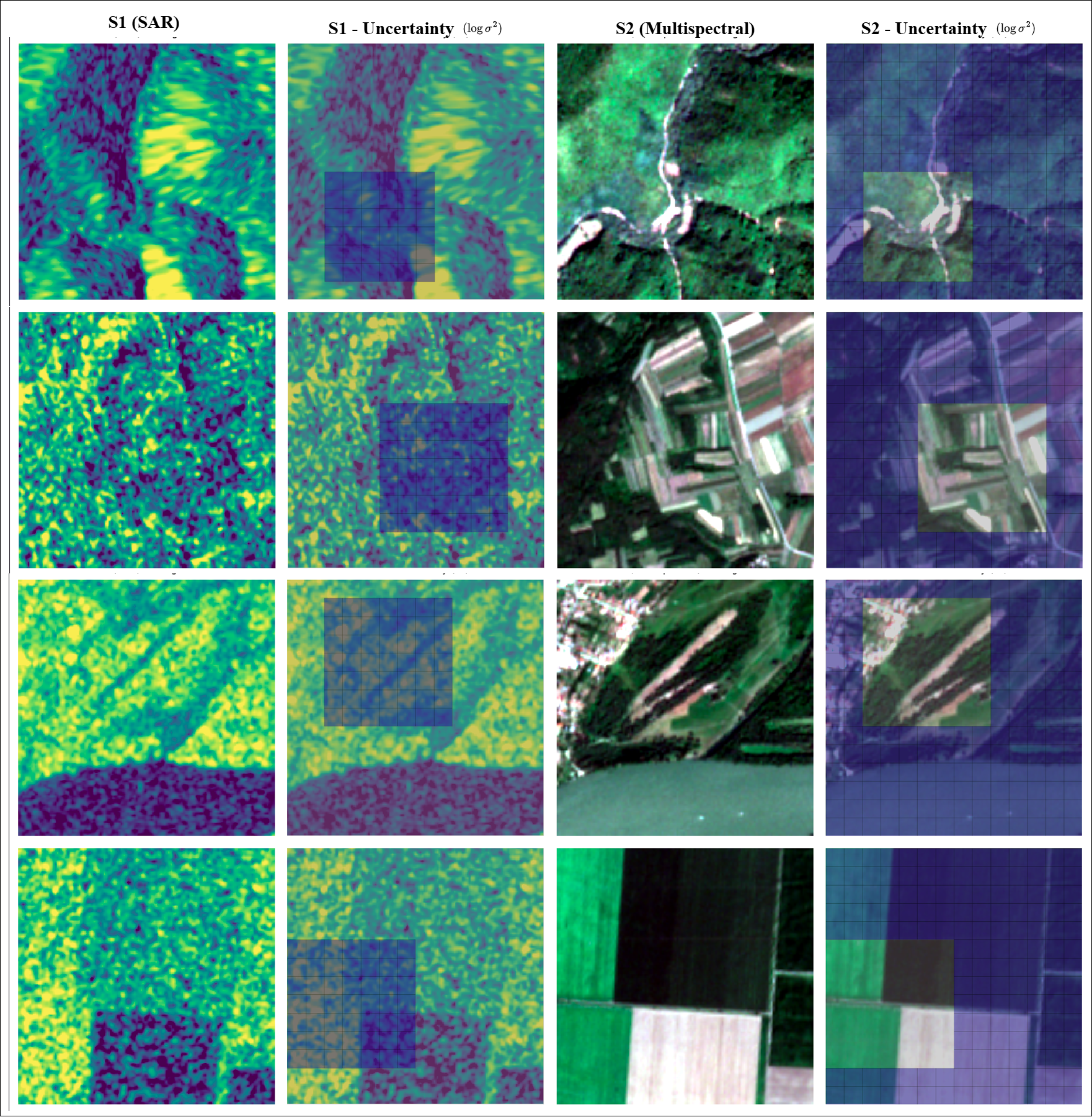}
\caption{\textbf{Per-patch predictive uncertainty} ($\log\sigma^2$) for four
S1--S2 scene pairs (rows: mountain/river, agricultural, 
grasslands/urban, agricultural). Uncertainty is shown as a boxed 
overlay on the masked region for both S1 SAR (col.~2) and S2 
multispectral (col.~4).}

\label{fig:uncertainty}
\end{figure*}

\vspace{-10pt}
\subsection{Comparisons to the Literature}

Table~\ref{tab:main} reports retrieval performance across all eight directions
on BEN-14K and fMoW. MIM-based methods struggle in cross-modal settings
despite strong unimodal performance: MAE achieves only 41.78 F1@5 on
S1$\to$S2, confirming that reconstructing SAR speckle and multispectral
spectral profiles in pixel space actively hinders modality-invariant
representation learning. CSMAE-SESD collapses specifically on S2$\to$S2 (39.01) despite
reasonable S1$\to$S1 performance (70.62), highlighting that pixel-space
reconstruction of multispectral spectral profiles overfits to low-level
spectral texture cues that do not transfer to semantic similarity. Cross-modal SSL methods (CROMA, SkySense, DeCUR) improve by
learning jointly across sensors, yet remain limited by false-negative
sensitivity in semantically dense RS scenes and the absence of global
distributional alignment. REJEPA demonstrates the advantage of latent
prediction over pixel reconstruction, while X-JEPA establishes a strong
cross-modal JEPA baseline (62.74, 62.98 on BEN-14K cross-modal directions).

CAPE-JEPA achieves state-of-the-art on all eight directions. On BEN-14K, it
surpasses X-JEPA by \textbf{6.91}, \textbf{3.40}, \textbf{13.74}, and
\textbf{13.52} F1@5 points on S1$\to$S1, S2$\to$S2, S1$\to$S2, and
S2$\to$S1 respectively. The cross-modal gains are particularly significant
in the RS context: SAR and multispectral sensors capture fundamentally
different physical phenomena, making cross-modal retrieval the hardest
direction and the one where deterministic predictors fail most severely due
to irreducible SAR speckle uncertainty. On fMoW, CAPE-JEPA surpasses X-JEPA
by \textbf{4.17}, \textbf{1.98}, \textbf{8.16}, and \textbf{15.59} F1@5
points on RGB$\to$RGB, SEN$\to$SEN, RGB$\to$SEN, and SEN$\to$RGB
respectively. Notably, CAPE-JEPA outperforms SkySense~\cite{guo2024skysense}
(398.04M) and CROMA~\cite{fuller2023croma} (310.54M), both pretrained on
substantially larger corpora, while using fewer parameters than X-JEPA
(407.30M vs.\ 432.72M). t-SNE and modality gap visualizations confirming 
distributional alignment are provided in  \textbf{Section~6 of the supplementary} 
(Fr\'{e}chet distance: CAPE-JEPA \textbf{10.8} vs.\ X-JEPA \textbf{22.9}).



Figure~\ref{fig:qualitative} shows top-5 retrievals across four cross-modal
directions against X-JEPA~\cite{Choudhury_2026_WACV},
SkySense~\cite{guo2024skysense}, and CROMA~\cite{fuller2023croma}. CAPE-JEPA
consistently retrieves semantically coherent scenes despite stark visual
differences across modalities, while baselines retrieve fewer relevant images.
Failure cases involve spectrally ambiguous land-cover boundaries where multiple
semantic classes co-exist at the patch scale.

In terms of \textbf{predictive uncertainty} estimation, 
Figure~\ref{fig:uncertainty} visualises the per-patch log-variance
$\log\boldsymbol{\sigma}^2$ output by the CAPE-JEPA probabilistic
predictor for four S1--S2 scene pairs. Context patches are shown as 
boxed overlays on the masked region for which no prediction is made. 
Two consistent patterns emerge across samples. First, \emph{high 
uncertainty} concentrates at semantically ambiguous regions: 
land-cover boundaries where multiple classes co-exist at the patch 
scale, SAR speckle-dominated areas where microwave backscatter is 
inherently stochastic, and urban--vegetation transitions where spectral 
signatures vary sharply within a single patch. Second, \emph{low 
uncertainty} characterises semantically homogeneous regions such as 
uniform cropland and open water, where the opposite modality provides 
strong predictive constraints. This spatial structure confirms that 
the Gaussian NLL loss~\cite{kendall2017uncertainties,sluijterman2023optimal} 
does not degenerate to uniform variance but learns to allocate 
predictive confidence in proportion to cross-modal ambiguity, using 
the variance signal to down-weight ambiguous patches during training 
and preventing the overconfident embeddings that arise when a 
deterministic predictor~\cite{Choudhury_2026_WACV} treats all patches 
as equally predictable. Additional uncertainty visualizations covering diverse 
land cover types are provided in \textbf{Section~7} 
of the supplementary.


\vspace{-10pt}
\subsection{Ablation Analysis}

In each ablation, one design choice is varied while all other hyperparameters
are held at default values; rows with suboptimal configurations are therefore
not directly comparable to the main comparison table (Table~\ref{tab:main}).

\noindent\textbf{Component Ablation (Table~\ref{tab:ablation_components}).}
The base model using only $\mathcal{L}_{\text{NLL}}$ already achieves
competitive cross-modal performance (72.25, 71.16 on S1$\to$S2 and S2$\to$S1),
confirming that cross-attention over opposite-modality context provides a strong
inductive bias. Adding SIGReg~\cite{balestriero2025lejepa,epps1983test} provides
consistent gains primarily on unimodal directions
(S1$\to$S1: +3.61, S2$\to$S2: +1.58), by stabilising each
modality's embedding space independently, with marginal gains
on cross-modal directions. Critically, neither XSIGReg nor InfoNCE is effective
in isolation: XSIGReg without InfoNCE causes cross-modal performance degradation
(S2$\to$S1: $-$2.03) as distributional alignment collapses inter-class structure
without discriminative grounding; InfoNCE without XSIGReg collapses S2$\to$S1
by 8.89 points as contrastive clusters resist cross-modal geometric alignment.
Only the full objective achieves optimal performance; Consistent mAP trends are in \textbf{Table~1 of the supplementary}.
\vspace{-8pt}
\begin{table}[H]
\centering
\setlength{\abovecaptionskip}{2pt}
\setlength{\belowcaptionskip}{2pt}
\caption{Component ablation on BEN-14K (F1@5, \%).}
\label{tab:ablation_components}
\vspace{2pt}
\setlength{\tabcolsep}{5pt}
\renewcommand{\arraystretch}{1.15}
\resizebox{0.5\columnwidth}{!}{%
\begin{tabular}{ccc cccc}
\toprule
\rowcolor{gray!15}
\textbf{SIGReg} & \textbf{XSIGReg} & \textbf{NCE}
& S1$\to$S1 & S2$\to$S2 & S1$\to$S2 & S2$\to$S1 \\
\midrule
\rowcolor{maerow!40}
\xmark & \xmark & \xmark & 70.42 & 65.12 & 72.25 & 71.16 \\
\rowcolor{maerow!20}
\cmark & \xmark & \xmark & 74.03 & 66.70 & 72.35 & 71.50 \\
\rowcolor{crossmodal!30}
\cmark & \cmark & \xmark & 75.24 & 68.41 & 74.05 & 74.47 \\
\rowcolor{crossmodal!30}
\cmark & \xmark & \cmark & 73.50 & 70.44 & 72.83 & 67.61 \\
\rowcolor{ourrow!60}
\cmark & \cmark & \cmark
& \textbf{79.22} & \textbf{80.64} & \textbf{76.48} & \textbf{76.50} \\
\bottomrule
\end{tabular}%
}
\end{table}
\vspace{-6pt}

\noindent\textbf{Probabilistic vs.\ Deterministic Prediction.}
CAPE-JEPA~(MSE) denotes a deterministic variant that replaces $\mathcal{L}_{\text{NLL}}$ with MSE loss. Table~\ref{tab:main} shows gains of \textbf{4.53} --\textbf{10.79} F1@5 from probabilistic 
modeling across all cross-modal directions.

\noindent\textbf{Masking Strategy (Table~\ref{tab:ablation_masking}).}
Complementary masking outperforms independent masking on three of four
directions, with a dramatic +8.05 gain on S2$\to$S2, by forcing each predictor
to exploit opposite-modality features. The single exception
(S2$\to$S1) reflects that independent masking preserves more local S2 context,
benefiting SAR retrieval when S2 features are already discriminative. Consistent mAP trends are in \textbf{Table~2 of the supplementary}.
\vspace{-8pt}
\begin{table}[H]
\centering
\setlength{\abovecaptionskip}{2pt}
\setlength{\belowcaptionskip}{2pt}
\caption{Masking strategy ablation on BEN-14K (F1@5, \%).}
\label{tab:ablation_masking}
\vspace{2pt}
\setlength{\tabcolsep}{5pt}
\renewcommand{\arraystretch}{1.15}
\resizebox{0.5\columnwidth}{!}{%
\begin{tabular}{l cccc}
\toprule
\rowcolor{gray!15}
\textbf{Strategy} & S1$\to$S1 & S2$\to$S2 & S1$\to$S2 & S2$\to$S1 \\
\midrule
\rowcolor{maerow!40}
Independent   & 66.81 & 72.12 & 73.17 & \textbf{75.94} \\
\rowcolor{ourrow!60}
Complementary & \textbf{79.04} & \textbf{80.17} & \textbf{75.16} & 74.04 \\
\bottomrule
\end{tabular}%
}
\end{table}
\vspace{-8pt}

\noindent\textbf{Masking Ratio (Table~\ref{tab:ablation_ratio}).}
Lower ratios (0.15--0.25) yield better unimodal performance while higher ratios
(0.40--0.75) improve S2$\to$S1 with significant unimodal degradation
($-$5.17 on S1$\to$S1),confirming aggressive masking trades representation quality for cross-modal reliance. Consistent mAP trends for masking ratio are in \textbf{Table~3 of the supplementary}.

\noindent\textbf{Backbone Scaling (Table~\ref{tab:ablation_backbone}).}
Scaling from ViT-S to ViT-L does not yield monotonic improvement: ViT-B is
consistently best for cross-modal retrieval (S2$\to$S1: 76.35), while ViT-L
underperforms on three of four directions, suggesting BEN-14K is too small to
saturate the larger backbone. ViT-S outperforms ViT-B on S1$\to$S1 (76.15
vs.\ 75.86), suggesting smaller backbones generalise better on the noisier
SAR modality at this data scale. Note that backbone ablation numbers differ from Table~\ref{tab:main}, which uses the full configuration including all three objectives and post-pretraining alignment,
whereas this ablation varies only the backbone. Consistent mAP trends are in \textbf{Table~4 of the supplementary}.

\vspace{-6pt}
\begin{table}[H]
\centering
\setlength{\abovecaptionskip}{2pt}
\setlength{\belowcaptionskip}{2pt}
\begin{minipage}{0.48\columnwidth}
\centering
\caption{Masking ratio ablation on BEN-14K (F1@5, \%).}
\label{tab:ablation_ratio}
\vspace{4pt}
\setlength{\tabcolsep}{4pt}
\renewcommand{\arraystretch}{1.15}
\resizebox{0.9\linewidth}{!}{%
\begin{tabular}{l cccc}
\toprule
\rowcolor{gray!15}
\textbf{Ratio} & S1$\to$S1 & S2$\to$S2 & S1$\to$S2 & S2$\to$S1 \\
\midrule
\rowcolor{jeparow!50}
0.15--0.25 & \textbf{78.68} & \textbf{72.44} & 71.21 & 71.37 \\
\rowcolor{jeparow!30}
0.30--0.50 & 77.37 & 69.67 & \textbf{72.05} & 73.46 \\
\rowcolor{maerow!40}
0.40--0.75 & 73.51 & 67.76 & 69.78 & \textbf{76.15} \\
\bottomrule
\end{tabular}%
}
\end{minipage}
\hfill
\begin{minipage}{0.48\columnwidth}
\centering
\caption{Backbone scaling ablation on BEN-14K (F1@5, \%).}
\label{tab:ablation_backbone}
\vspace{4pt}
\setlength{\tabcolsep}{4pt}
\renewcommand{\arraystretch}{1.15}
\resizebox{0.9\linewidth}{!}{%
\begin{tabular}{l c cccc}
\toprule
\rowcolor{gray!15}
\textbf{Backbone} & \textbf{Params} & S1$\to$S1 & S2$\to$S2
& S1$\to$S2 & S2$\to$S1 \\
\midrule
\rowcolor{crossrow!40}
ViT-S & 104.2M & 76.15 & 77.80 & 73.81 & 66.92 \\
\rowcolor{ourrow!60}
ViT-B & 407.3M & 75.86 & \textbf{79.92} & \textbf{74.53} & \textbf{76.35} \\
\rowcolor{maerow!40}
ViT-L & 928.4M & \textbf{78.29} & 76.95 & 73.15 & 73.35 \\
\bottomrule
\end{tabular}%
}
\end{minipage}
\end{table}
\vspace{-6pt}

\noindent\textbf{Alignment Epochs (Table~\ref{tab:ablation_align}).}
S1$\to$S1, S2$\to$S2, and S2$\to$S1 peak at 25 epochs. Extended alignment
overfits to unimodal discriminative structure, catastrophically degrading
cross-modal geometry (S2$\to$S1 drops from 76.18 at 25 to 66.92 at 100
epochs). We use 25 alignment epochs in the final model. This post-pretraining
alignment stage is integral to the CAPE-JEPA training pipeline
and is not applicable to baselines, which do not employ a predictive
pretraining objective. Minor differences from the main results
(Table~\ref{tab:main}) reflect that this ablation uses a fixed
configuration rather than the optimal hyperparameter combination.
Consistent mAP trends for alignment epochs are in
\textbf{Table~5 of the supplementary material}.

\noindent\textbf{Regularization Weight (Table~\ref{tab:ablation_sigreg_weight}).}
$\lambda_1$ controls the strength of isotropic Gaussian conditioning. Values
too small allow dimensional collapse and anisotropy; values too large
over-regularize, suppressing discriminative structure. $\lambda_1{=}0.01$
achieves the best cross-modal performance on both directions. Consistent mAP trends are in \textbf{Table~8 of the supplementary}.

\vspace{-6pt}
\begin{table}[H]
\centering
\setlength{\abovecaptionskip}{2pt}
\setlength{\belowcaptionskip}{2pt}
\begin{minipage}{0.48\columnwidth}
\centering
\caption{Alignment epochs on BEN-14K (F1@5, \%).}
\label{tab:ablation_align}
\vspace{2pt}
\setlength{\tabcolsep}{4pt}
\renewcommand{\arraystretch}{1.15}
\resizebox{0.9\linewidth}{!}{%
\begin{tabular}{c cccc}
\toprule
\rowcolor{gray!15}
\textbf{Epochs} & S1$\to$S1 & S2$\to$S2 & S1$\to$S2 & S2$\to$S1 \\
\midrule
\rowcolor{maerow!40}
0   & 70.17 & 75.70 & 73.57 & 66.02 \\
\rowcolor{ourrow!60}
25  & \textbf{79.22} & \textbf{79.82} & 75.06 & \textbf{76.18} \\
\rowcolor{jeparow!40}
50  & 75.10 & 76.90 & \textbf{76.48} & 67.49 \\
\rowcolor{maerow!40}
100 & 76.15 & 75.43 & 73.81 & 66.92 \\
\bottomrule
\end{tabular}%
}
\end{minipage}
\hfill
\begin{minipage}{0.48\columnwidth}
\centering
\caption{Regularization weight $\lambda_1$ on BEN-14K (F1@5, \%).}
\label{tab:ablation_sigreg_weight}
\vspace{2pt}
\setlength{\tabcolsep}{4pt}
\renewcommand{\arraystretch}{1.15}
\resizebox{0.9\linewidth}{!}{%
\begin{tabular}{c cccc}
\toprule
\rowcolor{gray!15}
$\lambda_1$ & S1$\to$S1 & S2$\to$S2 & S1$\to$S2 & S2$\to$S1 \\
\midrule
\rowcolor{maerow!40}
0.005 & 77.70 & 68.04 & 69.93 & 69.86 \\
\rowcolor{ourrow!60}
0.01  & \textbf{79.22} & \textbf{80.64} & \textbf{76.48} & 76.50 \\
\rowcolor{maerow!40}
0.05  & 66.29 & 69.65 & 73.18 & \textbf{76.64} \\
\bottomrule
\end{tabular}%
}
\end{minipage}
\end{table}
\vspace{-10pt}

\vspace{-10pt}
\section{Takeaways}
\label{sec:conclusion}
\vspace{-5pt}
We presented CAPE-JEPA, a probabilistic cross-modal Joint Embedding
Predictive Architecture for remote sensing image retrieval that addresses
three fundamental limitations of deterministic cross-modal prediction:
overconfident patch-level embeddings from deterministic prediction,
geometrically disjoint modality distributions, and semantic collapse
in the absence of discriminative grounding. By modelling cross-sensor prediction as a Gaussian distribution,
enforcing distributional alignment through adversarial characteristic
function tests, and anchoring scene-level discrimination via
in-training InfoNCE, CAPE-JEPA learns representations that are
simultaneously uncertainty-aware, geometrically well-conditioned,
and semantically discriminative. Evaluated on BEN-14K
and fMoW, CAPE-JEPA surpasses the strongest baseline X-JEPA by up to 
\textbf{13.74} F1@5 points on BEN-14K cross-modal retrieval 
and \textbf{15.59} points on fMoW, while using fewer parameters (407.30M vs.\ 432.72M). Ablation
studies confirm that probabilistic prediction, distributional alignment,
and discriminative grounding are complementary and each necessary for
optimal cross-modal performance.

{\small\noindent\textbf{Acknowledgements.}
This work was supported by the IIT 
Bombay-HRTI Centre of Excellence in Vision, 
Multimodal AI \& Geo Intelligence (VIMAGE) and the 
Anusandhan National Research Foundation (ANRF), 
Government of India, through the Core Research 
Grant (CRG\slash 2023\slash 004389).}
\clearpage

\bibliography{egbib}
\end{document}